\pdfoutput=1

\documentclass[11pt]{article}

\usepackage{acl}

\usepackage{times}
\usepackage{latexsym}
\usepackage{adjustbox}

\usepackage[T1]{fontenc}

\usepackage[utf8]{inputenc}

\usepackage{microtype}

\newcommand{\ie}{{\textit{i.e.~}}}

\title{Adversarial Text Normalization}

\author{Joanna Bitton \\ Meta AI \\  \small\texttt{jbitton@fb.com}
   \\\And
  Maya Pavlova \\ University of Waterloo \\ \small\texttt{mspavlova@uwaterloo.ca}
   \\  \\\And
  Ivan Evtimov \\ Meta AI \\ \small\texttt{ivanevtimov@fb.com}
  }

\begin{document}
\maketitle
\begin{abstract}
Text-based adversarial attacks are becoming more commonplace and accessible to general internet users. As these attacks proliferate, the need to address the gap in model robustness becomes imminent. While retraining on adversarial data may increase performance, there remains an additional class of character-level attacks on which these models falter. Additionally, the process to retrain a model is time and resource intensive, creating a need for a lightweight, reusable defense. In this work, we propose the \textit{Adversarial Text Normalizer}, a novel method that restores baseline performance on attacked content with low computational overhead. We evaluate the efficacy of the normalizer on two problem areas prone to adversarial attacks,~\ie Hate Speech and Natural Language Inference. We find that text normalization provides a task-agnostic defense against character-level attacks that can be implemented supplementary to adversarial retraining solutions, which are more suited for semantic alterations.

\end{abstract}

\section{Introduction}

\begin{figure}[t]
  \centering
  \includegraphics[width= \linewidth]{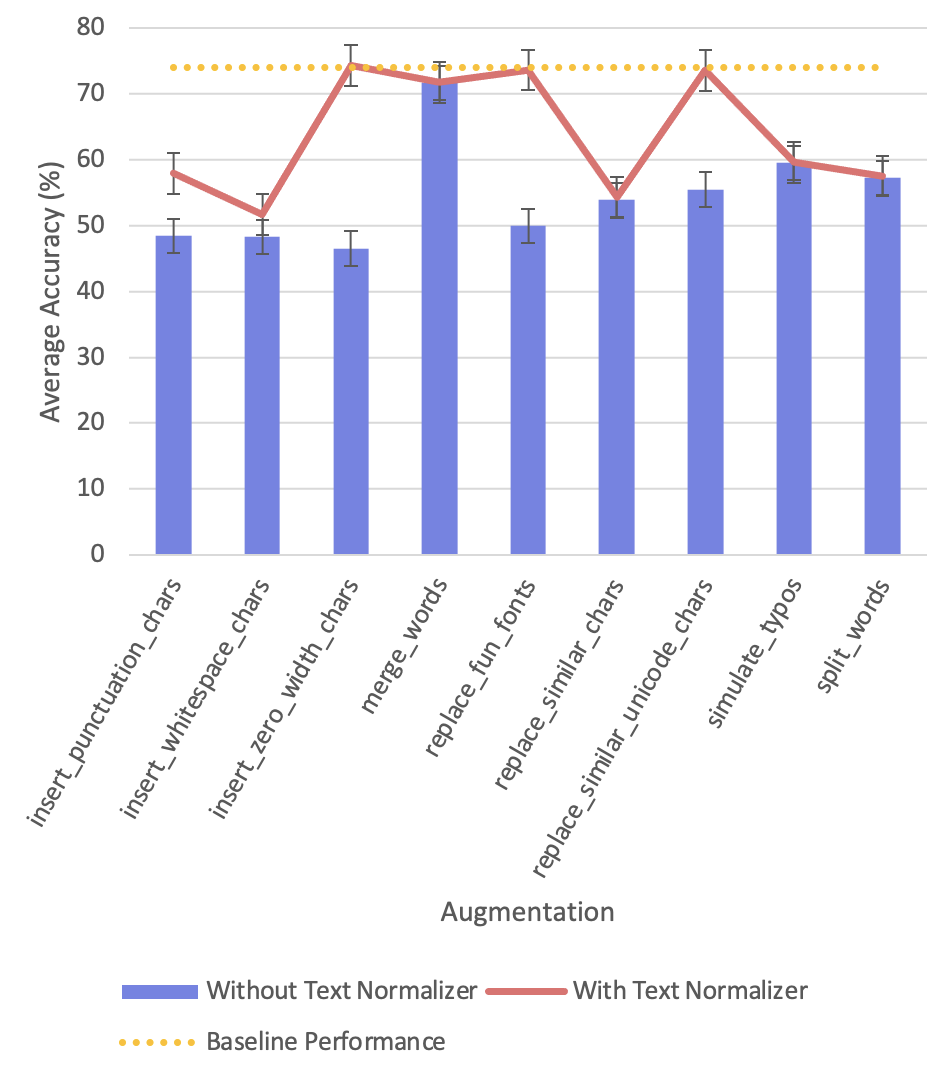}
  \caption{This figure showcases that text normalization is able to restore baseline scores on the Learning From The Worst (LFTW) test set for several augmentation types. The model scores in this graph are the averaged scores across all five LFTW models. To see the raw evaluation results, view Table \ref{tab:lftw-res} in the Appendix.}
  \label{fig:lftw}
\end{figure}

Natural language processing (NLP) models help preserve the integrity of discourse in online social networks by detecting hate speech, misinformation, and other content that violates community policies~\cite{halevy2020preserving}.
In these application scenarios, classifiers operate under significantly more adversarial conditions than in the standard paradigm of model development. 
Users often post content that is heavily altered in order to induce worst-case errors, dramatically reducing model performance relative to a standard test set. 
Recent research in machine learning has made strides towards building robust models to defend against sophisticated adversaries. Nevertheless, our experience and experiments show that models remain vulnerable to many simple and intuitive attacks.

One such class of highly-effective attacks are \textit{syntactic attacks}, which include adding punctuation and spacing, replacing fonts, and inserting zero-width characters. 
Attackers commonly employ these methods with little or no expert knowledge of machine learning algorithms, and constitute a large proportion of practical threats to natural language processing models deployed in industry. 
In our experiments on state-of-the-art robust NLP models, such attacks can decrease a model's performance by more than half. 
Thus, these text-based attacks remain a serious and unaddressed problem for the at-scale deployment of language models in integrity areas.

The literature has proposed methods to improve adversarial robustness by retraining models on ``hard'' and adversarial examples~\cite{nie2020adversarial,vidgen2021learning} but we observe several open challenges with applying these approaches. 
First, as our experiments demonstrate, even state-of-the-art models trained on multiple iterations of adversarial data continue to lack robustness to easily accessible syntactic attacks.
For example, in Figure~\ref{fig:hate_check}, inserting zero-width characters reduces a hate speech classification model's accuracy to less than 15\%. 
Second, these approaches are not amenable to quick iteration.
Whenever a new adversarial attack becomes more prevalent, a model developer would need to collect new data or create synthetic examples of such attacks, retrain the model, and redeploy it. This costs time, computational resources and has an environmental impact~\citep{sustainableAI2021}.

To address these outstanding issues, we propose a novel lightweight and easily extensible method for recovering model performance in the face of adversaries who perform syntactic attacks.
Our key insight is that many adversarial modifications can be undone before they even reach the model with little computational overhead.
From a machine learning perspective, this can be thought of as restoring the distribution of inference-time inputs to the original distribution on which the model was trained.
From a computer security perspective, this method is analogous to sanitizing the inputs to programs so as to make the input safe for further processing. 
In this work, we present the design and implementation of a system called the \textit{Adversarial Text Normalizer} (ATN), that achieves this goal.
This iterative approach produces a defense mechanism that can be applied at scale in a lightweight fashion to ensure robust model performance. 

An important principle in computer security is that defense mechanisms should be evaluated against adaptive adversaries~\citep{Petitcolas2011}, \ie those that can adjust their techniques to actions by the defender.
Therefore, we also partner with red teaming experts skilled at creating novel adversarial inputs to classical computer systems to conduct an adaptive evaluation of the ATN.
Our text normalizer can provide sufficient robustness gains even in the face of such adaptive adversaries.

Our contributions are as follows:
\begin{itemize}
    \item We design and implement a system for undoing syntactic attacks on textual models called the Adversarial Text Normalizer.
    \item We conduct an adaptive attacker red teaming exercise to evaluate the ATN's performance against skilled human adversaries.
    \item Through extensive experiments on three different benchmarks, we evaluate the performance of the ATN and conclude that it successfully recovers the original performance of a model when faced with syntactic attacks.
\end{itemize}

\section{Related Work}
Several papers have introduced benchmarks for adversarial attacks on NLP systems. Attacks focused on preserving semantic content and grammar \cite{Jin_Jin_Zhou_Szolovits_2020, alzantot-etal-2018-generating, adv_ex_with_syntax_control} have been shown to be effective against state-of-the-art models at the cost of requiring a greater understanding of the sentence structure and task context. In contrast, \citet{eger-benz-2020-hero} propose a benchmark of character level, orthographic perturbations as more realistic attacks in general applications, attributing the success of their high performance attacks to large out-of-vocabulary rates and disruption to tokenization. Other work \citep{Eger2019TextPL,boucher2021bad} investigates the replacement of characters with visually similar embedding spaces and the insertion of zero-width characters, noting the effectiveness of those methods against NLP models but marginal effect on human legibility -- especially when perturbing key offensive words. For such targeted attacks, \cite{rodriguez2018shielding} use a simple string matching algorithm to filter obfuscated and negated key words \citep{galeano2017deobfusctaion}, focusing on a limited list of target vocables on each pass.

Our work focuses on the implementation of text normalization as a computationally inexpensive and reusable solution to mitigate a range of highly accessible but effective adversarial text attacks such as character insertions, replacements, and censorship. Concurrent work in the NLI domain has addressed the bias in model performance on classic test sets and adversarial user attacks through iterative human-and-model-in-the-loop \cite{nie2020adversarial} data generation and model training. Similarly, \citet{vidgen2021learning} proposed a complementary approach with the amalgamation of targeted annotator samples including challenging perturbations to generate adversarial data for hate speech classification. Both works explored leveraging domain-experienced annotator resources to progressively train more robust models with each successive iteration. Other works on small text perturbations such as adversarial typos, have proposed the use of robust token-level encodings \cite{jones-etal-2020-robust} and preceding word recognition models \cite{pruthi-etal-2019-combating} as reusable systems that are trained once and then reused across models and tasks. In this work we explore a more lightweight, systematic correction layer that does not require training to create model and task-agnostic defenses.

\section{Methodology}

\subsection{Models and Datasets}

We identify two natural language tasks with significant importance to industrial applications and adversarial pressure.
First, hate speech classification is the problem of detecting statements that are likely to cause harm and inject toxicity in online discourse.
It has now become standard practice for providers of services where people can post comments and discuss content to employ hate speech classification models. 
These models are set up as binary classification models that output a score for the ``hatefulness'' of a given input statement. 
Second, Natural Language Inference (NLI) has been adapted for the detection of misinformation~\cite{nie2020adversarial}. 
In this setting, NLI models aim to flag statements that do not receive support from reputable sources or directly contradict information in them. 
Thus, a model is given access to a set of support statements and a ``hypothesis'' and it outputs a 3-way classification from among ``supported,'' ``not supported,'' and ``not enough information.'' 
In the cases of ``supported'' or ``not supported,'' the model also outputs the statement that supports or refutes the hypothesis.

Since both tasks are the subject of adversarial pressure, there have been several proposed approaches to robustifying models trained on them.
Most notably, the Dynalab~\cite{vidgen2021learning} approach proactively samples ``hard'' examples by asking human raters to conceive inputs that challenge the model.
Researchers then retrain the models and repeat the process for several rounds.
This paradigm helps achieve a large increase in robustness through the rounds, so we evaluate our approaches on those models as the benchmark for state-of-the-art robust performance.

For Hate Speech, we utilize the HateCheck~\citep{rottger2020ds} dataset (2,563 examples) and the Learning from the Worst (LFTW)~\citep{vidgen2021learning} test set (4,120 examples). The performance of state-of-the-art models trained on adversarial data from Learning from the Worst (LFTW) were compared with and without the addition of the text normalizer on both baseline and augmented versions of the dataset. Additionally, we chose to evaluate NLI models on the test sets from all three rounds of Adversarial NLI~\citep{nie2020adversarial} (1,000, 1,000, and 1,200 examples respectively). We assessed the performance of these baseline, augmented, and normalized datasets on five model architectures trained on SNLI~\citep{snli:emnlp2015}, MNLI~\citep{williams2018}, FEVER~\citep{Thorne18Fever} and all three rounds of Adversarial NLI. To see which models we evaluate on, please review Table \ref{table:models}. We specifically choose models already trained on adversarial datasets to assess opportunities for improvement beyond retraining. 

\begin{table}
\centering
\begin{tabular}{lll}
\hline
\textbf{Problem Area} & \textbf{Model} & \textbf{\# Parameters}\\
\hline
Hate Speech & LFTW & 125,000,000\\
Adversarial NLI & DeBERTa & 140,000,000\\
Adversarial NLI & RoBERTa & 125,000,000\\
Adversarial NLI & T5 & 220,000,000\\
Adversarial NLI & BERT & 109,000,000\\
Adversarial NLI & ALBERT & 17,000,000\\
\hline
\end{tabular}

\caption{\label{table:models} The models, associated problem areas, and number of parameters used in our evaluations. All LFTW models, from rounds 1-4 and more, have an identical number of parameters since they are all RoBERTa models.}
\end{table}

\subsection{Attacks}

\begin{table*}
\centering
\begin{tabular}{lll}
\hline
\textbf{Augmentation} & \textbf{Output Text} & \textbf{Normalized Text}\\
\hline
None & This is augmented text & This is augmented text\\
\small\texttt{insert\_punctuation\_chars} & Th.i.s ,is ...a.ug;m!en't?ed, ,te!x.t & This ,is ...augmented, ,text\\
\small\texttt{insert\_whitespace\_chars} & T h i s  is  a u g m e n t e d   text & This is augmented text\\
\small\texttt{insert\_zero\_width\_chars} & This is augmented text & This is augmented text\\
\small\texttt{merge\_words} & Thisis augmented text & Thisis augmented text \\
\small\texttt{replace\_fun\_fonts} & \includegraphics[height=0.48cm]{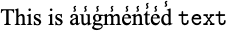} & This is augmented text\\
\small\texttt{replace\_similar\_chars} & Th!s is @ugmented tex7 & Th!s is @ugmented tex7\\
\small\texttt{replace\_similar\_unicode\_chars} & \includegraphics[height=0.43cm]{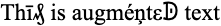} & This is augmented text \\
\small\texttt{simulate\_typos} & This is augmentde texht & This is augmentde texht\\
\small\texttt{split\_words} & Th is is augment ed text & Th is is augment ed text \\
\hline
\end{tabular}
\caption{\label{table:text-augs} Examples of augmentations generated using the open-source library AugLy~\citep{papakipos2022augly}, leveraged in the analysis for adversarial attacks on the text datasets, and their normalized counterparts. Note that while the output of \texttt{insert\_zero\_width\_chars} appears visually identical to the original sentence, there are actually zero-width unicode characters embedded throughout the entire string. We include augmentations that are not covered by the normalizer (such as \texttt{merge\_words}) in our evaluations to showcase that our method does not further corrupt unknown attack types.}
\end{table*}

To attack the aforementioned datasets, we use the open-source augmentation library AugLy~\citep{papakipos2022augly} to simulate various text-based adversarial attacks commonly used on social media platforms. We focus on using character-level attacks as opposed to attacks that can potentially add, remove, or change full words in the text, to avoid perturbations in the semantic meaning. The specific augmentations selected for the analysis are listed in Table \ref{table:text-augs}.

In addition to the synthetically generated attacks, we collected 98 samples of hate speech text which were adversarially created and modified by individuals in a cybersecurity Red Team. In the first half of the session, participants were tasked with creating their own attacks based off of their prior knowledge of text-based attacks seen online. In the second half of the session, they were given direct access to the code implementation for the text normalizer and were tasked to bypass it using targeted attacks. 

Many of the attacks created were similar to the attacks of \texttt{insert\_punctuation\_chars}, \texttt{replace\_fun\_fonts}, \texttt{simulate\_typos}, \texttt{replace\_similar\_unicode\_chars}, and \texttt{replace\_similar\_chars}. The adversaries also created letter repetition attacks, i.e. ``hellllooooo", censored violating text, and replaced words with emojis.

\subsection{Adversarial Text Normalizer}

The text normalizer is an isolated correction unit that can be placed in front of models to target character-level attacks for various NLP tasks. The normalizer is designed to be used as a preprocessing step prior to text tokenization. For optimal computational efficiency, the operator is written in torchscript, and can process approximately 77 examples per second on a server with an Intel(R) Xeon(R) CPU E5-2680 v4 @ 2.40GHz processor. This operator was incorporated into the PyTorch model as a customizable data transform. The algorithm relies on sophisticated string manipulation as a targeted defense against known adversarial attacks, and is easily scaled up to support additional attacks as the adversarial environment progresses. 

Currently, the text normalizer supports removing three overarching categories of text attacks:

\begin{enumerate}
    \item \textbf{Character insertion}: addition of characters such as punctuation marks, whitespaces, Unicode characters, emojis, and more to separate characters in a word with the intent to disrupt proper tokenization. This category includes augmentation methods such as \texttt{insert\_punctuation\_chars}, \texttt{insert\_whitespace\_chars}, and \texttt{insert\_zero\_width\_chars}.
    \item \textbf{Character replacement}: substitution of standard Latin characters with visually similar characters from other languages or Unicode characters with the intent of obfuscation. This category includes augmentation methods such as \texttt{replace\_fun\_fonts}, \texttt{replace\_similar\_chars}, and \texttt{replace\_similar\_unicode\_chars}.
    \item \textbf{Censorship of violating words}: replacement of letters in violating words with punctuation characters to avoid explicit content. For instance ``kill" could be censored as ``k***", ``k!ll", ``k\#*!" and more.
\end{enumerate}

To undo the effects of a character insertion attack such as \texttt{insert\_punctuation\_chars}, the text is first split by whitespaces to identify `words'. For each word, we then determine how many extraneous punctuation characters have been inserted, ignoring punctuation marks at the beginning and end, as such additions do not segment the word to disrupt tokenization. If the amount of punctuation characters is below a set threshold or the word resembles a URL, we do not modify the word and add it to our normalized string. Otherwise, we replace the superfluous punctuation with spaces, strip the string of excessive whitespace, and concatenate consecutive single character entities together.

For character replacement attacks, we predefined multiple mappings between Unicode characters and their keyboard character pairs, and performed a string search method to reverse the replacement.

As for censorship attacks, a list of common user-censored toxic terms were identified prior based on flagged user content. For each toxic term, a regex for the censorship pattern was defined such that the first and last letters of the word remain constant but any of the letters in between can be replaced by punctuation characters - maintaining the same length as the original, uncensored word. For every match found, we replace the censored string with its uncensored pair accordingly.

\section{Evaluation}

In this section, we examine the performance of Hate Speech and Natural Language Inference (NLI) models with respect to the original, the augmented, and normalized datasets to evaluate the efficacy of the text normalizer. All evaluations were conducted using Dynabench~\citep{dynaboard2021}, in which AWS ECR models are deployed as endpoints and Batch Transform jobs are run on AWS Sagemaker to get dataset predictions. We roughly spent 38.36 CPU hours on model inference (no GPUs were used). 

\subsection{Hate Speech}

\begin{figure}[t]
  \centering
  \includegraphics[width= \linewidth]{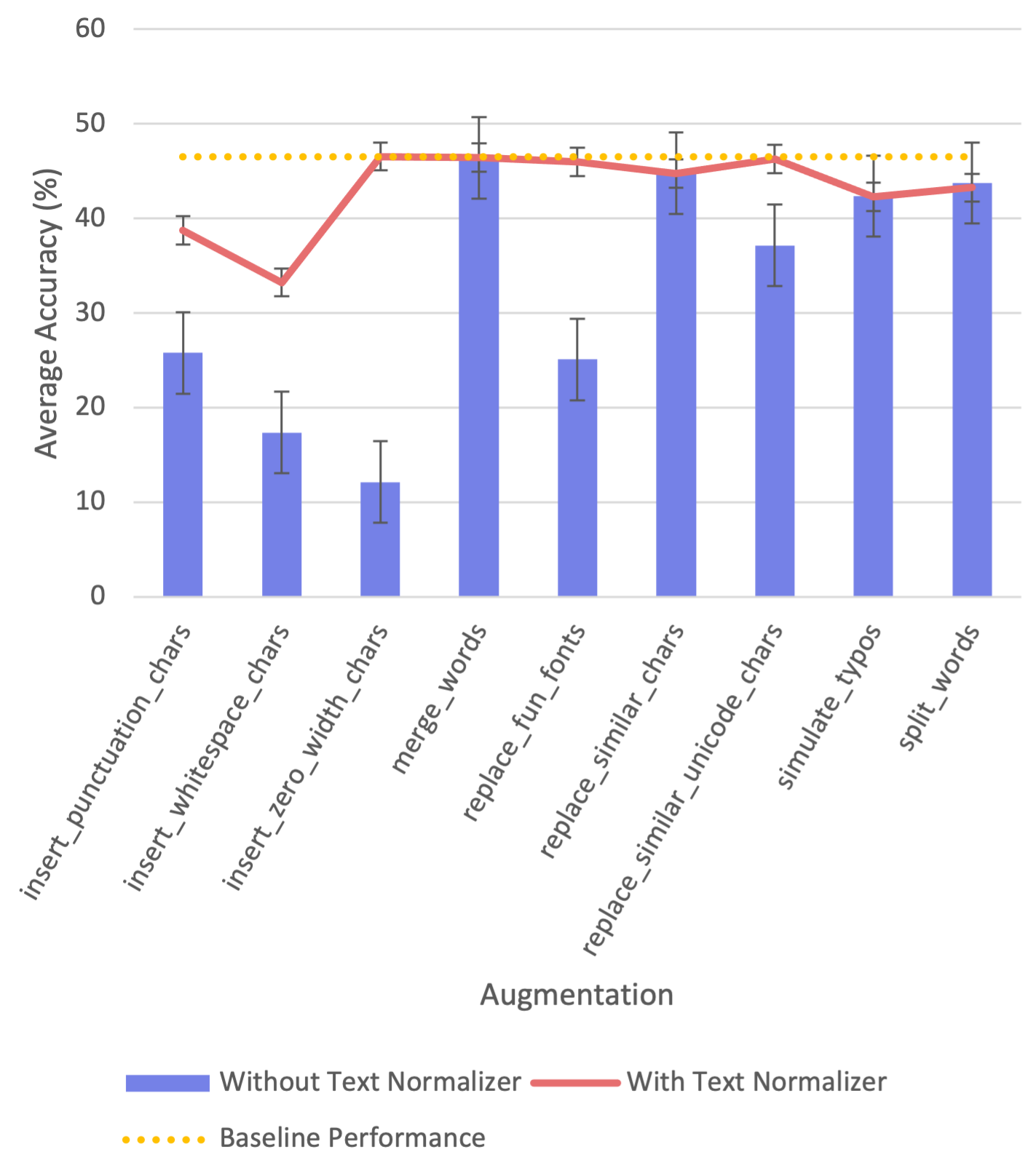}
  \caption{This figure showcases the results of evaluating Hate Speech models against the baseline, augmented, and normalized HateCheck datasets. The model scores in this graph are the averaged scores across all five LFTW models. To see the raw evaluation results, view Table \ref{tab:hc-res} in the Appendix.}
  \label{fig:hate_check}
\end{figure}

To assess performance on adversarial hate speech data, we evaluated on five models from the Learning from the Worst (LFTW)~\citep{vidgen2021learning} paper that were previously retrained on varying amounts of "rounds" of adversarial hate speech data collection.

\subsubsection{Learning from the Worst}

We first evaluate on the test set from Learning from the Worst. Figure \ref{fig:lftw} showcases these results. Overall, the text normalizer maintains or improves initial performance on all augmented datasets and the baseline. Across all models, normalizing the \texttt{insert\_zero\_width\_chars}, \texttt{replace\_similar\_unicode\_chars}, and \texttt{replace\_fun\_fonts} augmentations resulted in the most significant performance gains, with a maximum of a \textbf{32.18\%} increase. In between, normalizing \texttt{insert\_punctuation\_chars} and \texttt{insert\_whitespace\_chars} had increases of at most 11.95\%. For the LFTW R3 model, normalizing the whitespace text resulted in a 1.27\% loss in performance. This may be due to the fact that not every whitespace character in the AugLy augmentation is removed by the normalizer. As expected, other augmentations that aren't covered by the normalizer did not see any substantial gains or losses in performance. To view the raw model scores, see Table \ref{tab:lftw-res} in the Appendix.

\subsubsection{HateCheck}

In addition to evaluating on the LFTW test set, we also evaluated on an out-of-distribution dataset, HateCheck. Figure \ref{fig:hate_check} displays these results. The trends observed in the LFTW test set overall have agreement with the HateCheck results. However, in this case, there were no losses in performance for normalized augmentations covered by our method. The largest performance gain was \textbf{48.89\%} by normalizing \texttt{insert\_zero\_width\_unicode\_chars}, and the smallest performance gain was 6.1\% by normalizing \texttt{insert\_punctuation\_chars}. To see the raw evaluation results, please review Table \ref{tab:hc-res} in the Appendix.

\subsubsection{Red Team Attacks}

Lastly, we evaluated the LFTW models on the Red Team dataset tasked to bypass the text normalizer. The baseline scores across all models averaged at 39.52\% and the normalized scores averaged at 41.32\%. In comparison to the synthetically-augmented text, applying the normalizer resulted in a less drastic increase in performance. This difference can be attributed to the fact that a significant amount of the data was attacked similarly to \texttt{replace\_similar\_chars} and \texttt{simulate\_typos}, two augmentations not covered by our defenses. To view the raw model scores, see Table \ref{tab:red-team-res} in the Appendix.

\subsection{Natural Language Inference}

To validate our results in another problem space, we evaluated on five Natural Language Inference (NLI) models previously trained on a collection of adversarial and benign NLI datasets. We evaluated the models on the baseline, augmented, and normalized ANLI test sets from rounds 1-3.

\begin{figure}[t]
  \centering
  \includegraphics[width= \linewidth]{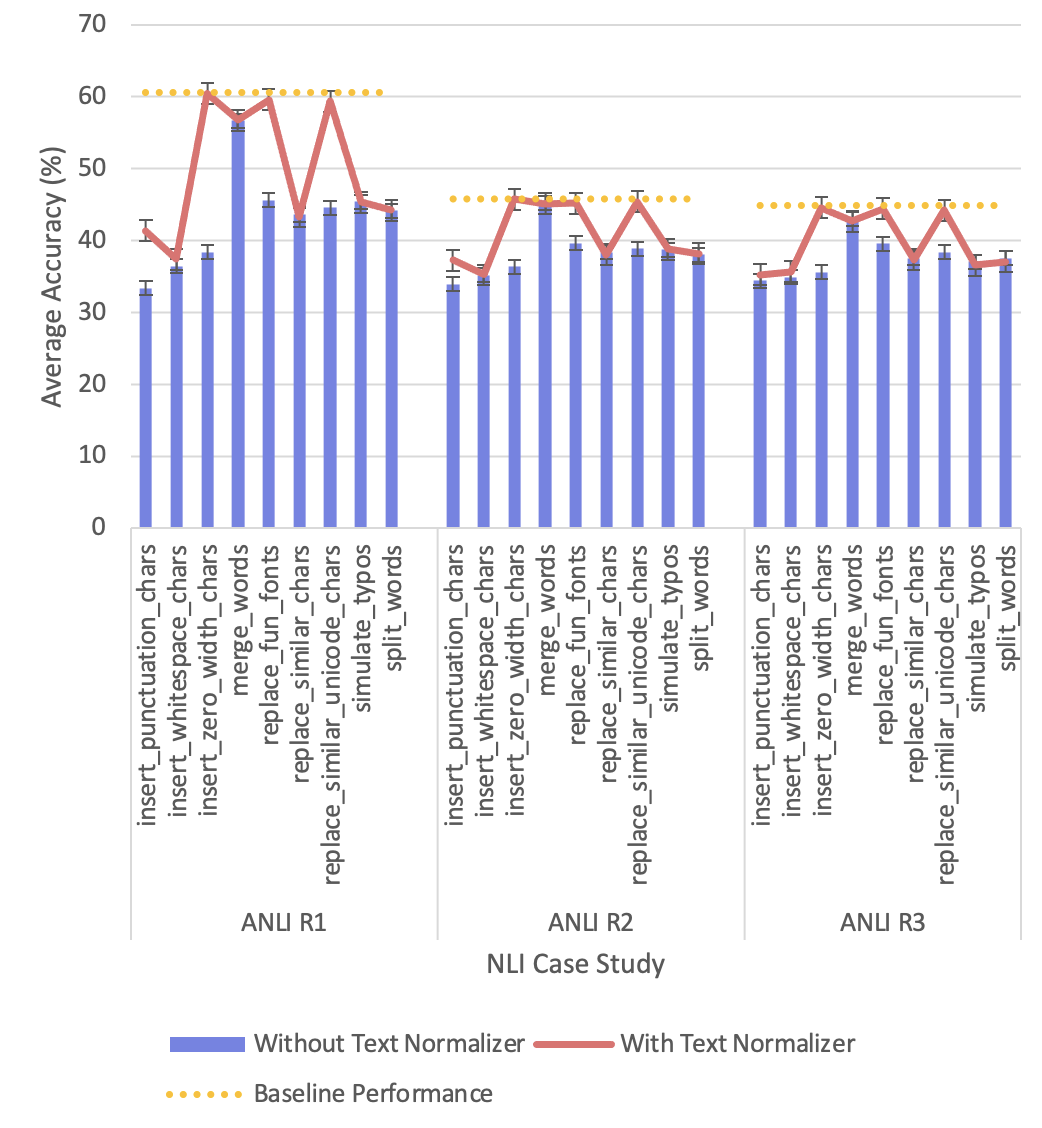}
  \caption{This figure contains the results of evaluating ANLI models against the baseline, augmented, and normalized rounds 1-3 test sets. The model scores in this graph are the averaged scores across all five models. To see the raw evaluation results, view Tables \ref{tab:anlir1-res}, \ref{tab:anlir2-res}, and \ref{tab:anlir3-res} in the Appendix.}
  \label{fig:anli}
\end{figure}

Overall, the results from the ANLI experiments align with the previous insights discussed in the Hate Speech task. The
\texttt{replace\_fun\_fonts}, \texttt{insert\_zero\_width\_chars}, and \texttt{replace\_similar\_unicode\_chars} attacks were the most performant, followed by \texttt{insert\_punctuation\_chars}, \texttt{insert\_whitespace\_chars}, and finally the non-covered attack types. The largest gain, \textbf{29.1\%}, was made by normalizing \texttt{insert\_zero\_width\_unicode\_chars} in the ANLI R1 test set. To view the raw model scores, see Tables \ref{tab:anlir1-res}, \ref{tab:anlir2-res}, and \ref{tab:anlir3-res} in the Appendix.


\section{Discussions}

Retraining models on adversarial data has been proposed as a mitigation method for adversarial attacks~\citep{vidgen2021learning,nie2020adversarial}. Its benefits are that it's relatively simple to implement, shown to be effective~\citep{goodfellow2015explaining}, and doesn't affect model throughput once deployed. However, if a new adversarial attack were to be discovered on a system, the turnaround time for deploying a robust model can be quite long. A developer would need to (1) gather and annotate or systematically generate the attacked data (2) retrain the model (3) redeploy the model. Steps (1) and (2) could be highly nontrivial depending on the attack type and model size. In addition, there can also be a significant monetary and environmental cost in retraining a large model~\citep{sustainableAI2021}.

On the other hand, text normalization allows a developer to move fast. Instead of collecting data, an engineer would simply need to write one additional function to reverse said attack and test it. In addition, it's lightweight and there is no need to retrain, as this is a text preprocessing step. However, text normalization cannot handle every attack type. Text normalization is best suited to mitigate character-level attacks that do not change the semantic meaning of the text, \ie~syntactic attacks. 

Another important consideration is that text normalization \textit{does} affect real-time performance. A balance must be found between intelligently mitigating attacks and compute. In our use case, we limited the normalizer to sophisticated string manipulation, as anything more would result in too much compute. Hence, it becomes less feasible to build defenses that require knowledge or understanding of words, etc.

Thus, we recommend the best way to mitigate adversarial attacks is to use a combination of text normalization and retraining. Specifically, retraining should be used for semantic adversarial attacks, and adversarial text normalization for syntactic adversarial attacks. Despite all the models evaluated on being retrained on adversarial data, they were still vulnerable to character-level text attacks. However, together with the text normalizer, we were able to increase performance and even return to baseline performance at times. This study is meant to show that text normalization in general is a viable approach to mitigating syntactic text-based attacks. This is certainly not the final method, and we hope researchers extend this work to support multilingual text.

\section{Conclusion}

We proposed a new method to mitigate text-based adversarial attacks, called the Adversarial Text Normalizer. We evaluated the performance of models retrained on adversarial data with and without the ATN.
Our experiments show that text normalization and retraining should be used together in order to maintain baseline performance against a broad range of adversaries.

\section*{Acknowledgements}
The authors would like to thank Adina Williams, Tristan Thrush, and Kushal Tirumala for their support in running experiments with the Dynabench platform. 
The authors would also like to thank Luke Zettlemoyer, Caner Hazirbas, and Stefan Hermanek for helpful discussions on the work in this paper.
Finally, the authors extend their gratitude to Aaron Grattafiori, Tom Ravenscroft, Adi Ajit, Aimee Couglin, Athena Cheung, Ben Actis, Christopher Korban, Greg Prosser, Hamza Kwisaba, James Burton, Jayson Grace, Martin Vigo, Nat Hirsch, Ryan Hall, Sasha Hanganu, Tom Hebb, and Vlad Ionescu for participating in the evaluation of the Adversarial Text Normalizer.

\section*{Impact Statement}
\subsection*{Risks from the Work}
Our work is meant to reduce risk to online discourse by enabling NLP models to function robustly in an adversarial setting.
We acknowledge that releasing this work might enable stronger attacks against some systems, but we believe it is important to discuss these defenses publicly for two reasons.
First, other practitioners and researchers can benefit from our findings in building stronger defenses.
Second, the history of security and cryptographic research clearly demonstrates that robust systems are built only when they go through ongoing attack/defend iteration cycles. 
To that end, we hope that our work informs the next steps in building robust NLP systems.
As with any NLP system and computer technology, we acknowledge robustness may be at odds with safety when the models defended with our mechanism are themselves used for nefarious purposes. 

\subsection*{Use of Scientific Artifacts}
In this work, we made heavy use of the Dynabench platform~\citep{dynabench2021,dynaboard2021} and the models trained on data collected with it. 
We worked closely with the creators of the platform and the models, and they were always fully aware of our intentions.
The Dynabench platform has been released under the MIT license: \url{https://github.com/facebookresearch/dynabench/blob/main/LICENSE}.
We also used the Adversarial NLI, HateCheck, and Learning from the Worst datasets. 
Each of those do not contain identifying information and only associate a pseudonymous annotator ID with each example.
We verified this manually by looking at samples from the datasets.
The HateCheck and Learning from the Worst datasets contain offensive language as part of the nature of the task they were set up for.
All datasets are in English and were created by English-speaking authors and annotators in the United States.

\subsection*{Humans Involved in the Research}
We did not employ annotators or perform human subject experiments.
We engaged with a partner team in a collegial capacity to expand our insights into the adversarial text normalizer and we discuss those findings here for the benefit of the broader research community.



\bibliographystyle{acl_natbib}
\bibliography{anthology}

\newpage
\appendix
\section*{Appendix}
\section{AugLy Augmentation Generation}

To augment the hate speech and natural language inference datasets, we chose random parameters for each augmentation for every piece of text. The ranges we used for every parameter are listed below. Since multiple AugLy augmentations accept as input the same parameters, we do not break this down by augmentation type (the same value was used across all augmentations). The following ranges were inputted into \texttt{random.uniform()} and lists of options were inputted \texttt{random.choice()}:

\begin{itemize}
    \item \textbf{\texttt{aug\_p}}: (0.3, 1.0)
    \item \textbf{\texttt{aug\_word\_p}}: (0.3, 1.0)
    \item \textbf{\texttt{aug\_char\_p}}: (0.1, 0.4)
    \item \textbf{\texttt{granularity}}: ["char", "word", "all"]
    \item \textbf{\texttt{vary\_fonts}}: [True, False]
\end{itemize}

\section{Raw Data: Hate Speech and Natural Language Inference Analyses}
\label{sec:appendix-raw-tables}
Here, we provide the full evaluation results with exact numbers broken down by attack type and model.

\begin{table}
\centering
\begin{tabular}{lcc}
\hline
\textbf{Model} & \textbf{Red Team} & \textbf{Normalized} \\
\hline
LFTW R1 & 32.41 & 35.95\\
LFTW R2 & 40.61 & 41.32\\
LFTW R3 & 39.88 & 41.32\\
LFTW R4 & 42.01	& 43.68\\
All LFTW & 42.69 & 44.32\\
\hline
\end{tabular}
\caption{Hate Speech Case Study Results by Model: Red Team Dataset}
\label{tab:red-team-res}
\end{table}

\begin{table*}
\centering
\begin{adjustbox}{width=\columnwidth*2,center}
\begin{tabular}{lcccccc}
\hline
\textbf{Augmentation} & \textbf{LFTW R1} & \textbf{LFTW R2} & \textbf{LFTW R3} & \textbf{LFTW R4} & \textbf{All LFTW} \\
\hline
\texttt{baseline} & 57.79 & 70.03 & 81.20 & 81.09 & 80.27\\
\texttt{normalized} & 58.04 & 70.08 & 81.24 & 81.21 & 80.27\\
\hline
\texttt{insert\_punctuation\_chars} & 43.34 & 50.48 & 48.53 & 49.23 & 50.74\\
\texttt{normalized} & 50.83 & 58.48 & 60.48 & 63.40 & 56.43\\
\hline
\texttt{insert\_whitespace\_chars} & 43.05 & 51.97 & 53.65 & 46.89 & 45.84\\
\texttt{normalized} & 48.83 & 52.97 & 52.38 & 52.50 & 52.05\\
\hline
\texttt{insert\_zero\_width\_chars} & 35.04 & 50.73 & 51.04 & 48.96 & 46.87\\
\texttt{normalized} & 58.37 & 70.23 & 81.29 & 81.14 & 80.38\\
\hline
\texttt{merge\_words} & 56.47 & 68.74 & 77.67 & 78.27 & 77.30\\
\texttt{normalized} & 56.67 & 68.83 & 77.73 & 78.30 & 77.10\\
\hline
\texttt{replace\_fun\_fonts} & 42.23 & 50.08 & 55.75 & 51.10 & 50.43\\
\texttt{normalized} & 58.11 & 70.67 & 80.42 & 80.01 & 78.91\\
\hline
\texttt{replace\_similar\_chars} & 52.43 & 56.21 & 52.39 & 52.36 & 56.17\\
\texttt{normalized} & 52.48 & 56.76 & 53.00 & 52.76 & 56.19\\
\hline
\texttt{replace\_similar\_unicode\_chars} & 47.65 & 56.16 & 56.26 & 58.06 & 59.34\\
\texttt{normalized} & 58.20 & 70.15 & 80.25 & 80.04 & 79.08\\
\hline
\texttt{simulate\_typos} & 53.95 & 60.48 & 61.61 & 59.93 & 61.81\\
\texttt{normalized} & 53.99 & 60.55 & 61.34 & 60.31 & 61.74\\
\hline
\texttt{split\_words} & 52.45 & 55.74 & 59.27 & 58.36 & 60.22\\
\texttt{normalized} & 52.38 & 56.38 & 59.32 & 58.77 & 60.82\\
\hline
\end{tabular}
\end{adjustbox}
\caption{Hate Speech Case Study Results: LFTW}
\label{tab:lftw-res}
\end{table*}

\begin{table*}
\centering
\begin{adjustbox}{width=\columnwidth*2,center}
\begin{tabular}{lcccccc}
\hline
\textbf{Augmentation} & \textbf{LFTW R1} & \textbf{LFTW R2} & \textbf{LFTW R3} & \textbf{LFTW R4} & \textbf{All LFTW} \\
\hline
\texttt{baseline} & 36.48 & 47.78 & 49.23 & 49.35 & 49.45\\
\texttt{normalized} & 36.45 & 47.73 & 49.29 & 49.37 & 49.47\\
\hline
\texttt{insert\_punctuation\_chars} & 13.18 & 35.05 & 32.52 & 30.09 & 18.09\\
\texttt{normalized} & 28.19 & 41.15 & 40.78 & 44.67 & 38.95\\
\hline
\texttt{insert\_whitespace\_chars} & 11.19 & 15.72 & 29.63 & 15.41 & 14.96\\ 
\texttt{normalized} & 20.60 & 34.48 & 45.57 & 33.01 & 32.57\\
\hline
\texttt{insert\_zero\_width\_chars} & 0.00 & 21.28 & 30.50 & 8.37 & 0.58\\ 
\texttt{normalized} & 36.45 & 47.73 & 49.29 & 49.37 & 49.47\\
\hline
\texttt{merge\_words} & 37.03 & 47.99 & 48.72 & 49.16 & 49.15\\
\texttt{normalized} & 37.32 & 47.92 & 48.75 & 49.17 & 49.08\\
\hline
\texttt{replace\_fun\_fonts} & 16.38 & 27.04 & 32.68 & 27.43 & 21.95\\ 
\texttt{normalized} & 36.12 & 46.32 & 48.83 & 49.29 & 49.38\\
\hline
\texttt{replace\_similar\_chars} & 33.86 & 47.63 & 47.57 & 47.77 & 47.09\\
\texttt{normalized} & 34.10 & 47.47 & 47.46 & 47.80 & 47.05\\
\hline
\texttt{replace\_similar\_unicode\_chars} & 22.61 & 39.35 & 41.07 & 42.68 & 40.06\\
\texttt{normalized} & 36.42 & 47.26 & 49.12 & 49.29 & 49.45\\
\hline
\texttt{simulate\_typos} & 32.46 & 44.16 & 42.84 & 46.49 & 45.88\\
\texttt{normalized} & 32.57 & 43.81 & 42.79 & 46.44 & 45.75\\
\hline
\texttt{split\_words} & 33.76 & 47.19 & 44.11 & 47.21 & 46.49\\
\texttt{normalized} & 33.20 & 46.44 & 43.58 & 47.00 & 46.04\\
\hline
\end{tabular}
\end{adjustbox}
\caption{Hate Speech Case Study Results: HateCheck}
\label{tab:hc-res}
\end{table*}

\begin{table*}
\centering
\begin{adjustbox}{width=\columnwidth*2,center}
\begin{tabular}{lcccccc}
\hline
\textbf{Augmentation}  & \textbf{RoBERTa} & \textbf{T5} & \textbf{BERT} & \textbf{ALBERT} & \textbf{DeBERTa} \\
\hline
\texttt{baseline} & 62.10 & 58.90 & 53.80 & 63.00 & 65.00 \\
\texttt{normalized} & 61.80 & 58.90 & 54.00 & 62.90 & 65.00 \\
\hline
\texttt{insert\_punctuation\_chars} & 32.40 & 32.70 & 32.60 & 34.40 & 34.60 \\
\texttt{normalized} & 40.40 & 38.80 & 41.20 & 39.90 & 46.40 \\
\hline
\texttt{insert\_whitespace\_chars} & 36.40 & 38.50 & 40.20 & 33.60 & 33.20 \\
\texttt{normalized} & 38.00 & 39.80 & 39.50 & 34.60 & 35.10 \\
\hline
\texttt{insert\_zero\_width\_chars} & 35.50 & 34.30 & 53.80 & 31.90 & 36.20 \\
\texttt{normalized} & 61.80 & 58.70 & 54.00 & 62.80 & 65.30 \\
\hline
\texttt{merge\_words} & 60.10 & 57.20 & 48.00 & 58.30 & 59.90 \\
\texttt{normalized} & 60.00 & 57.30 & 48.10 & 58.50 & 59.70 \\
\hline
\texttt{replace\_fun\_fonts} & 40.00 & 50.10 & 36.40 & 62.30 & 39.10 \\
\texttt{normalized} & 60.50 & 58.50 & 51.80 & 63.10 & 64.00 \\
\hline
\texttt{replace\_similar\_chars} & 45.30 & 44.20 & 39.60 & 44.00 & 44.70 \\
\texttt{normalized} & 43.80 & 43.80 & 40.50 & 43.80 & 44.60 \\
\hline
\texttt{replace\_similar\_unicode\_chars} & 44.30 & 42.10 & 38.20 & 53.40 & 44.90 \\
\texttt{normalized} & 61.00 & 58.00 & 52.40 & 62.00 & 63.60 \\
\hline
\texttt{simulate\_typos} & 46.70 & 44.20 & 42.90 & 45.00 & 48.10 \\
\texttt{normalized} & 46.80 & 44.20 & 42.90 & 44.90 & 47.70 \\
\hline
\texttt{split\_words} & 45.50 & 42.90 & 41.40 & 44.30 & 46.60 \\
\texttt{normalized} & 45.80 & 42.80 & 41.60 & 43.70 & 47.10 \\
\hline
\end{tabular}
\end{adjustbox}
\caption{Natural Language Inference Case Study Results: ANLI R1}
\label{tab:anlir1-res}
\end{table*}

\begin{table*}
\centering
\begin{adjustbox}{width=\columnwidth*2,center}
\begin{tabular}{lcccccc}
\hline
\textbf{Augmentation} & \textbf{RoBERTa} & \textbf{T5} & \textbf{BERT} & \textbf{ALBERT} & \textbf{DeBERTa} \\
\hline
\texttt{baseline} & 46.50 & 46.80 & 44.80 & 46.50 & 44.50\\
\texttt{normalized} & 46.20 & 46.80 & 44.90 & 46.60 & 44.40 \\
\hline
\texttt{insert\_punctuation\_chars} & 35.40 & 34.10 & 31.80 & 34.20 & 34.10 \\
\texttt{normalized} & 35.60 & 36.10 & 37.90 & 38.50 & 38.00 \\
\hline
\texttt{insert\_whitespace\_chars} & 35.30 & 36.70 & 35.50 & 33.90 & 34.40 \\
\texttt{normalized} & 33.60 & 36.50 & 35.80 & 34.80 & 35.30 \\
\hline
\texttt{insert\_zero\_width\_chars} & 35.10 & 34.20 & 44.80 & 33.40 & 34.10 \\
\texttt{normalized} & 46.30 & 46.70 & 44.90 & 46.60 & 44.30 \\
\hline
\texttt{merge\_words} & 45.60 & 46.60 & 46.70 & 44.30 & 42.80 \\
\texttt{normalized} & 45.50 & 46.60 & 46.60 & 44.50 & 42.50 \\
\hline
\texttt{replace\_fun\_fonts} & 36.70 & 42.20 & 35.90 & 46.40 & 36.80 \\
\texttt{normalized} & 44.80 & 45.90 & 44.80 & 46.30 & 44.00 \\
\hline
\texttt{replace\_similar\_chars} & 38.00 & 37.90 & 37.50 & 39.60 & 38.80 \\
\texttt{normalized} & 38.20 & 37.60 & 36.20 & 39.40 & 38.60 \\
\hline
\texttt{replace\_similar\_unicode\_chars} & 39.40 & 38.30 & 36.90 & 42.00 & 37.60 \\
\texttt{normalized} & 46.00 & 46.20 & 44.30 & 46.20 & 44.40 \\
\hline
\texttt{simulate\_typos} & 38.40 & 39.10 & 39.00 & 37.70 & 39.40 \\
\texttt{normalized} & 38.30 & 38.90 & 39.30 & 37.70 & 39.70 \\
\hline
\texttt{split\_words} & 38.00 & 38.30 & 36.80 & 37.40 & 39.50 \\
\texttt{normalized} & 38.30 & 38.20 & 36.10 & 38.40 & 39.80 \\
\hline
\end{tabular}
\end{adjustbox}
\caption{Natural Language Inference Case Study Results: ANLI R2}
\label{tab:anlir2-res}
\end{table*}

\begin{table*}
\centering
\begin{adjustbox}{width=\columnwidth*2,center}
\begin{tabular}{lcccccc}
\hline
\textbf{Augmentation} & \textbf{RoBERTa} & \textbf{T5} & \textbf{BERT} & \textbf{ALBERT} & \textbf{DeBERTa} \\
\hline
\texttt{baseline} & 45.58 & 44.83 & 44.00 & 44.17 & 45.83 \\
\texttt{normalized} & 45.08 & 44.83 & 43.67 & 43.58 & 45.33 \\
\hline
\texttt{insert\_punctuation\_chars} & 35.17 & 33.50 & 34.58 & 33.67 & 35.17 \\
\texttt{normalized} & 35.75 & 36.58 & 33.92 & 33.92 & 36.00 \\
\hline
\texttt{insert\_whitespace\_chars} & 33.67 & 35.17 & 36.33 & 34.33 & 34.92 \\
\texttt{normalized} & 34.75 & 36.25 & 37.33 & 35.00 & 35.00 \\
\hline
\texttt{insert\_zero\_width\_chars} & 34.42 & 32.25 & 44.00 & 32.08 & 35.00 \\
\texttt{normalized} & 45.08 & 44.83 & 43.83 & 43.75 & 45.33 \\
\hline
\texttt{merge\_words} & 43.42 & 44.58 & 41.42 & 40.33 & 44.92 \\
\texttt{normalized} & 42.83 & 44.58 & 41.42 & 40.00 & 44.42 \\
\hline
\texttt{replace\_fun\_fonts} & 38.08 & 40.92 & 37.08 & 44.25 & 37.50 \\
\texttt{normalized} & 45.33 & 44.92 & 43.58 & 43.58 & 44.83 \\
\hline
\texttt{replace\_similar\_chars} & 40.00 & 37.08 & 36.92 & 36.83 & 37.00 \\
\texttt{normalized} & 38.83 & 37.42 & 36.42 & 36.75 & 37.17 \\
\hline
\texttt{replace\_similar\_unicode\_chars} & 38.67 & 38.00 & 36.17 & 41.17 & 37.67 \\
\texttt{normalized} & 44.58 & 45.00 & 43.08 & 43.58 & 44.67 \\
\hline
\texttt{simulate\_typos} & 35.92 & 36.83 & 36.83 & 37.75 & 38.00 \\
\texttt{normalized} & 35.50 & 36.67 & 36.00 & 37.00 & 37.50 \\
\hline
\texttt{split\_words} & 38.50 & 38.50 & 37.08 & 37.75 & 35.67 \\
\texttt{normalized} & 37.50 & 38.17 & 36.42 & 37.33 & 35.75 \\
\hline
\end{tabular}
\end{adjustbox}
\caption{Natural Language Inference Case Study Results: ANLI R3}
\label{tab:anlir3-res}
\end{table*}

\end{document}